\documentclass[conference]{IEEEtran}
\usepackage{cite}
\ifCLASSINFOpdf
 \usepackage[pdftex]{graphicx}
 \graphicspath{{./images/}}
\else
\fi
\usepackage[cmex10]{amsmath}
\usepackage{mathtools}
\usepackage{textgreek}
\renewcommand\IEEEkeywordsname{Keywords}

\begin{document}
\title{Preprocessing Methods of Lane Detection and Tracking for Autonomous Driving}

\author{\IEEEauthorblockN{Akram Heidarizadeh}
\IEEEauthorblockA{\textit{College of Eng., University of Tehran}\\
Tehran, Iran \\Email: a.heydarizadeh@ut.ac.ir}
}


%Telephone: (800) 555--1212, Fax: (888) 555--1212}


\maketitle

\begin{abstract}
In the past few years, researches on advanced driver assistance systems (ADASs) have been carried out and deployed in intelligent vehicles. Systems that have been developed can perform different tasks, such as lane keeping assistance (LKA), lane departure warning (LDW), lane change warning (LCW) and adaptive cruise control (ACC). Real time lane detection and tracking (LDT) is one of the most consequential parts to performing the above tasks.
Images which are extracted from the video, contain noise and other unwanted factors such as variation in lightening, shadow from nearby objects and etc. that requires robust preprocessing methods for lane marking detection and tracking.
Preprocessing is critical for the subsequent steps and real time performance because its main function is to remove the irrelevant image parts and enhance the feature of interest.
In this paper, we survey preprocessing methods for detecting lane marking as well as tracking lane boundaries in real time focusing on vision-based system.

\end{abstract}


% \fi
\IEEEpeerreviewmaketitle

\section{Introduction}
The increasing number of road accidents is a critical matter that modern society is confronted with. Driver inattention, fatigue and somnolence are the most common reasons that cause the accidents with high number of fatalities. Different measures are being applied by automotive industries in order to minimize the accidents by incorporating advanced driver assistant system in vehicles. Nowadays vehicles tend to be more intelligent, intelligent vehicles are the ones which apply various advanced technologies to capture real time information which is further analyzed to provide more comfort and safety for the drivers. Vision-based and Radar-based techniques are the most commonly used ones in intelligent vehicles. Vision-based techniques are recognized as robust and impressive methods used for advanced driver assistance system. Vision-based lane detection, traffic sign detection, pedestrian detection, vehicle detection, etc. are the most popular features that are extensively deployed in intelligent vehicles. 
Efforts have been done by several researchers all over the world for developing lane detection and tracking system.
Lane detection system faces many challenges and the robust techniques are needed to be developed to overcome these problems and to increase the efficiency of detection\cite{date2017vision}.\par
Most vision-based lane detection systems are commonly designed based on image processing techniques within similar frameworks. Vision-based lane detection systems described in studies usually consist of three main process, which are preprocessing, lane detection and lane tracking, as shown in Fig.1. Among these, the preprocessing process, which enhance the input image, is one of the most significant aspects of the real time lane detection system\cite{xing2018advances}.\par
A lane detection and tracking system meet too many challenges, as shown in Fig.2. Strange street materials and different slopes create problems in lane detection. 
Not only faded lanes on the road but also sharp and irregular shaped curves of lanes can reduce the quality of lane detection systems processing. It has been see that even the merged lanes have leaded to some detection problems.
Shadows made by trees on urban road are one of the most common conditions which can produce strength edges to mislead the lane detection. 
Environmental condition issues also affect on the image clarity like roads covered with the snow, fog, heavy rains, and reflection on wet roads with low visibility\cite{shirke2017study}.
On the other hand, in real time lane detection and tracking systems, it is very important to minimize the processed items to decrease the processing time. 
Proper implementation of the preprocessing part of a lane detection system is great importance in covering these challenges.
This study tries to comprehensively review the lane detection system with focusing on preprocessing block.\par
The rest of this paper is organized as below. In section II all preprocessing methods used in various papers have been exactly investigated. In section III and IV lane detection and lane tracking methods review respectively. Finally, the section V highlights the concluding remarks.
\begin{figure}[h]
\includegraphics[width=0.5\textwidth]{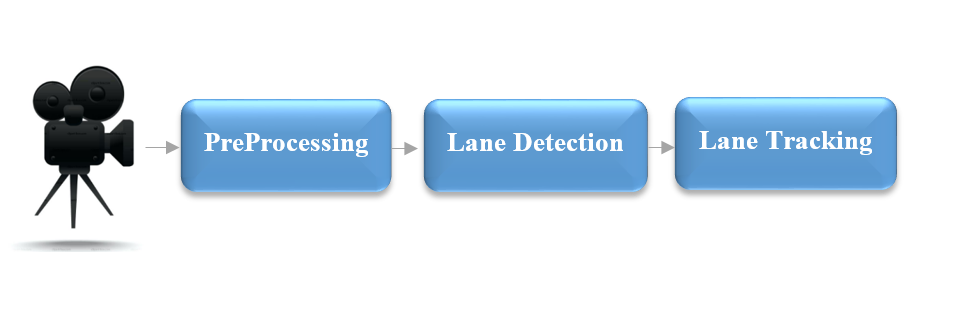}
\caption{General Framework of Lane Detection and Tracking Systems}
\centering
\end{figure}

\begin{figure}[h]
\includegraphics[width=0.5\textwidth]{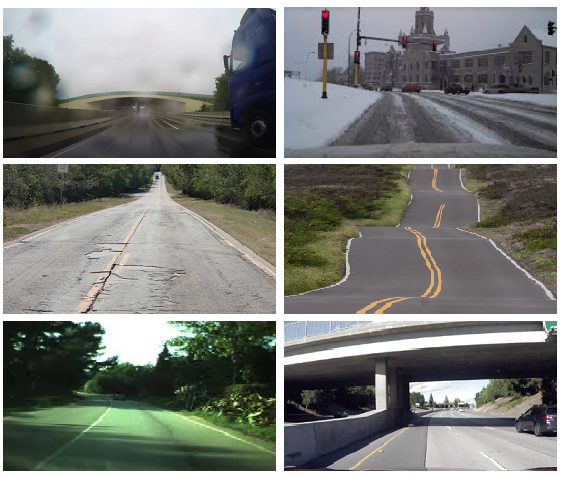}
\caption{A Number of Challenges to The Lane Detection and Tracking Systems}
\centering
\end{figure}

\section{Preprocessing}
As it mentioned, images which are extracted from the video have various noises.
The task of the preprocessing block is the removal of noise and other unwanted components of the image and the successful transfer of areas with useful information to the next stage.\par
The most common procedures in the preprocessing step include image smoothing, region of interest (ROI) selection, transferring color image into greyscale image or a different color format, inverse perspective mapping (IPM) also known as birds-eye view, segmentation and etc.
In the following, we carefully investigate all of the preprocessing steps.

\subsection{Image Smoothing}
In order to recognize lane markings, image smoothing can be performed using various filters including Median filter, Gaussian filter, Histogram filter, Dilation and Erosion filters, mean filter and etc. to blur the noisy details.\\

\subsubsection{Median filter}
In \cite{shihavuddin2008road}, \cite{truong2008new}, \cite{truong2008lane}, the median filter is used for smoothing. By using the median filter on the image, various noise can be deleted. In \cite{shihavuddin2008road}, the median filter is used to eliminate the heterogeneity of the road surface. In \cite{shihavuddin2008road}, for the pixel I (x.y), the median filter is applied as in equation (1).
\begin{equation}
I(x,y)=median\{I(x-k,y-1)|(k,1) \in w\}
\end{equation}
Where w is the chosen window. In \cite{zheng2008automatic}, a medium filter of $3\ast3$ was used for smoothing.

\subsubsection{Gaussian filter}
In \cite{yu2008lane}, to remove noise in the image, including noises caused by rain and snow and dirty roads, Equation (2) is used as a Gaussian pattern.
\begin{equation}
\frac{1}{16}\times
 \Bigg [\,
\begin{tabular}{ccc}
  1 & 2 & 1 \\
  2 & 4 & 2 \\
  1 & 2 & 1 
 \end{tabular}
 \Bigg ]\,
\end{equation}

With respect to the part of the image in the form of equation (3), the output of each pixel is obtained by equation (4).
\begin{equation}
A=  \Bigg ( \,
\begin{tabular}{ccc}
  a(m-1, n-1) & a(m, n-1) & a(m+1, n-1) \\
  a(m-1, n) & a(m, n) & a(m+1, n) \\
  a(m-1, n+1) & a(m, n+1) & a(m+1, n+1)
 \end{tabular}
 \Bigg ) \,
\end{equation}

\begin{equation}
a^\prime (m,n)=\frac{1}{4} (1\enspace2 \enspace 1)*A* \frac{1}{4} (1\enspace 2 \enspace1)^T  
\end{equation}

In \cite{hota2009simple} applied a Gaussian filter to smooth the input image based on gradient information.

\subsubsection{Histogram filter}
In \cite{bao2008lane}, the median filter and Gaussian filter have been used to reduce the random noise and Salt and Pepper noise. They then use the histogram filter to improve the brightness and contrast of the image. This action is performed using equation (5).

\begin{equation}
O (\,x,y ) \, =\frac{I ( \,x,y ) \, -min [ \,I ( \,x,y ) \, ] \,}{ max [ \,I ( \,x,y ) \, ] \,-min [ \,I ( \,x,y ) \, ] \,}*255
%
%max⁡[\,I ( \,x,y ) \,]\,- min⁡[\,I ( \,x,y ) \,]\,
\end{equation}

\subsubsection{Dilation and Erosion filters}
In \cite{he2003color}, in order to remove noise and simplify the computation in subsequent steps,  basic mathematical morphology operators, Dilation and Erosion filters, are used.
\subsubsection{Average filter}
In [13], a temporal blurring creates an average image $\overline{I}_K(n)$ to make the line markings appear long and continuous.
This smoothing helps connect dashed lane markers to form a near continuous line.
Consider a video clip that contains $N$ images where each image is indexed by $n$ and $n\in[1,N]$. The average image is created by computing the average between the current image $I(n)$ and $K$ images from the past. It is formed as follows:
\begin{equation}
\overline{I}_K(n)=\sum^{K}_{i=0}\frac{I(n-i.\bigtriangleup)}{K}
\end{equation}
where $i$ and $\triangle$ are used to create the offset from the current image index $n$.

\subsection{Region of Interest (ROI)}
The extraction of region of interest (ROI) is an important task of preprocessing stage, aiming at reducing the computational cost due to the processing time. It is unnecessary to process the entire pixels of images. Computation should be focused on regions which contain important information, as shown in Fig.3. To get the ROI, three main approaches can be found in literature, which are vanishing point detection, perspective analysis and projective model, and sub-sampling.

\begin{figure}[h]
\centering
\includegraphics[width=0.3\textwidth]{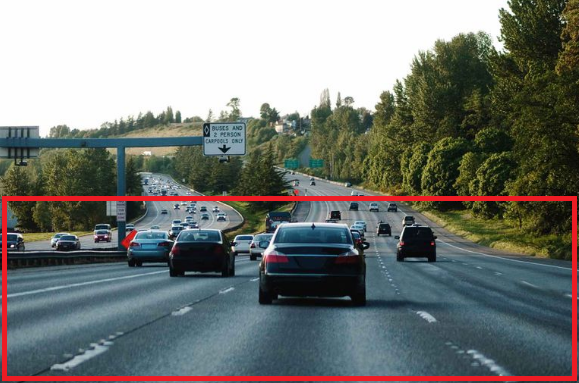}
\caption{Region of Interest}
\centering
\end{figure}

\subsubsection{Vanishing Point Extraction}
 A host of researchers utilized vanishing point for road following or as a directional constraint for defining the path boundaries. Vanishing point is defined as the point of intersection of the perspective projections of a set of parallel lines in 3D scene onto the image plane. For a straight road segment,vanishing point is obtained as the intersection point of the lines that characterize the lane, as shown in Fig.4(a). In the case of a curved road, vanishing point is approximated by the lane borders, markings etc. in the vicinity of the vehicle, as shown in Fig.4(b). Its distance components to the image center yield the relative direction (yaw and pitch angles) of the vehicle with respect to the path. Therefore, estimated vanishing point can be utilized to as a directional constraint for region of interest segmentation \cite{kong2009vanishing} or steer the vehicle \cite{rasmussen2008roadcompass}, lane marking detection \cite{hua2006fast} etc.

Over the past decades, vanishing point detection has been intensively studied. Some methods have been proposed to detect road vanishing point based on road edge \cite{matessi1999vanishing}. In this way, road boundaries or lane markings are first extracted by edge detector, and then straight lines are detected by Hough transform, from the intersections of line pairs, vanishing point can be detected. This kind of approaches may work well in structured roads with lane markings, however, these methods do not appropriately behave in unstructured roads with complex geometric characteristics.\par
In order to overcome the shortcomings of the above methods, texture-based methods have been proposed recently. The methods do not rely on road boundaries or lane markings, and their performances in unstructured roads are much better than that of edge-based methods.
Some researchers employed Gabor filters to estimate local textures orientations for the voting of road vanishing point \cite{rasmussen2008roadcompass}, \cite{kong2009vanishing}, \cite{moghadam2011fast}. However, the local dominant texture orientation in every pixel of the road image is estimated, and the vanishing point candidates are obtained in a greedy way. Thus, the computational cost is very high, so that these algorithms cannot be realized in real time.

\begin{figure}[h]
\centering
\includegraphics[width=0.5\textwidth]{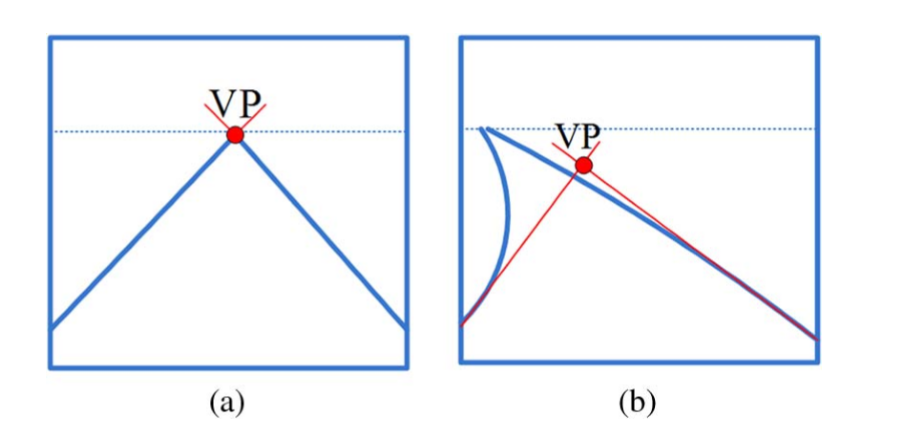}
\caption{Vanishing Point Extraction}
\centering
\end{figure}

\subsubsection{Perspective analysis and projective model}
The lane marking width and shape change as a result of perspective effect. Parallel lane markings in real world plane intersects at vanishing point in the image plane. Usually, by analysing the perspective effect, detection range can be focused on a certain area, which can be region of interest. With a reasonable projection applied between image plane, real world plane and camera plane, not only the region of interest can be extracted, it also benefits overcoming the perspective distortion, getting the position of vanishing point and bird’s eye view.\par
A projective model can be determined by computing the homography between real world plane and image plane.
Road plane needs to be transformed to image coordinate system through projection matrix given by camera calibration. It is noticeable that, as an essential component of a homography, the calibration matrix of camera is usually computed off line. As described in \cite{nieto2011road}, a pinhole camera projection model is used to compute the position of vanishing point, by computing the unknown parameters of the rotation matrix.\par
Rapidly Adapting Lateral Position Handler (RALPH)  constructed a very basic projection model to get region of interest, which is a trapezoid, to focus on lane marking areas, so that irrelevant area (road surface) can be eliminated. Due to the perspective effect, the bottom is shorter than the top of the trapezoid, with their length vary based on vehicle velocity \cite{pomerleau1995ralph}.

\subsubsection{Sub-Sampling}
It is also reasonable to split ROIs by sub-sampling. A predefined or adaptive percentage of the image can be used to determine the size of region of interest \cite{yu2008lane}. Besides, some researchers divide the image horizontally or vertically into small parts \cite{bellino2005lane}. By conducting different strategies for rural and urban areas, in \cite{nedevschi20043d} applied predefined percentage to split ROIs for rural ways and adaptive percentage for highways.

\subsection{Inverse Perspective Map (IPM)}
Inverse perspective mapping (IPM) is one of the most important parts of preprocessing that is commonly used. This may be realized by remapping each pixel towards a different position, usually birds’ eye view. Experimentally, it has already been proved that Lane detection within the image given by IPM performs much better than methods without IPM. Many researchers have used IPM to transform an image from a real world plane to a bird’s eye view \cite{deng2013real}, \cite{sivaraman2013integrated}, \cite{li2013sensor}.
As shown in Fig. 5, by eliminating the perspective effect, the desired lane marking candidates present in the form of straight and parallel lines in the remapped image.

\begin{figure}[h]
\centering
\includegraphics[width=0.5\textwidth]{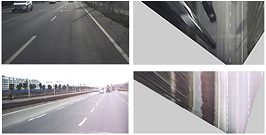}
\caption{Inverse perspective mapping}
\centering
\end{figure}

\subsection{Segmentation}
To prepare images for detection stage, segmentation can be accomplished by extracting certain features from the input image. Colour and edge are two main features which are considered for lane detection segmentation. The major problem with edge segmentation is that a number of thresholds need to be adjusted, and the adjustments depend only on the image undergoing testing, while colour segmentation techniques are sensitive to lightening conditions, especially those occurring on sunny days.

\subsubsection{Edge-based Segmentation}
Edge-based segmentation usually consists of edge detection and binarization. Different edge detection techniques are proposed based on edge features, which usually include shape (distribution of edge points) and gradient information. Canny algorithm is used in \cite{shen2013multi} and \cite{hota2009simple}. Steerable filter is another effective approach for edge segmentation. As demonstrated in \cite{mccall2006video}, edges are extracted based on lane orientation characteristics with applying steerable filter. Prweitt filter is another edge detector used in \cite{li2013sensor}.\par
A straightforward step after edge detection is to apply binarization, by doing this properly, effective details can be reserved while noisy points being neglected. Different methods have been proposed to find suitable threshold for binarization.
A predefined threshold based on massive experimental results is used for this task in \cite{shihavuddin2008road}. Because of the dependence on massive experimental results and previous experience, predefining the threshold can be inaccurate. Hence, adaptive threshold is proposed as a more effective approach for segmentation to binarize images \cite{wu2009dynamic}. 

\subsubsection{Colour-based Segmentation}
Most of the case, different colour features can be presented based on different colour spaces (RGB, HSV, gray scale, etc.). In the literature, data structure of the images can be changed in order for better performance based on different colour spaces. For structured roads (mostly highways), it is better to use gray images for colour representation of images to be detected, because gray images make lane markings with light colours (white or yellow) distinguished from road surface (with dark gray for most of the time). Hence, gray-level conversion is considered in many papers, such as \cite{satzoda2014drive} and \cite{nieto2011road}. As in \cite{nieto2011road}, a parametric multiple-class likelihood model of the road is proposed based on the gray level of pavement, lane markings and objects. Parameters are estimated by the Expectation-Maximization (EM) algorithm.\par
In \cite{li2004lane}, different from directly converting coloured images into gray level, only uses R and G channels of coloured images to form gray images. Furtherly, to make full use of the colour information of RGB images, some researchers deployed colour format conversion in their algorithm. For example, in \cite{lipski2008fast}, RGB images are converted to HSV. However, format conversion of coloured images may inevitably lose a certain amount of detail information in real scenarios. To solve this problem, in \cite{sun2006hsi} converted RGB to HSI. In in \cite{sun2006hsi} revealed the advantage of combining HSI model with loosen thresholds, in terms of keeping useful information and highlighting the difference between lane markings and road surface. The method proposed in \cite{wen2008road} is based on the colour consistency of the road surface. The road colour is assumed to be Gaussian distributed, by which the road surface colour and relevant variance can be computed.

\subsubsection{MSER-based Segmentation}
In \cite{sun2011automatic} proposed a method of describing MSER blobs using SIFT-based descriptors, followed by a graphical model in order to localize lane markings. This method combines unsupervised learning algorithm with off-line training. It consists of 3 main steps: 1) obtain descriptors by extracting the feature of lane markings; 2) quantize descriptors into which relies much on visual words retrieving and 3) the image is described as a combination of visual words.\\
There are two drawbacks for this method: 1) the processing time varies in different scenarios, as the number of features detected varies; 2) the visual words for the feature of lane markings mostly consist of line segments, which takes both salient lane markings and noisy lines into account. This might increase the false positive results.

\section{Lane Detection}

The second stage, detection, extracts lane markings from the output of the preprocessing step. In this section, lane detection methods are reviewed from the scope of conventional image processing and novel machine learning techniques. The first part will focus on the review of generally used conventional image processing methods. In the second part, lane detection algorithms based on machine learning and deep learning techniques will be discussed.

\subsection{Image Processing Based Lane Detection Algorithms}
This approach extracts lane markings by using feature extraction methods. Three main feature extraction approaches can be categorized in the literature: edge-based methods, colour-based methods and hybrid methods.

\subsubsection{Edge-Based Methods}
One of the most commonly edge extractors used in this task is Hough transform \cite{cualain2012automotive},\cite{deng2013real}.
However, Hough transform has some disadvantages such as its computational complexity and the unavoidably high false positive rate. Facing with this issue, variants of Hough Transform are used. Adaptive random Hough transform (ARHT)\cite{li2004lane} and probabilistic Hough transform (PHT) \cite{hota2009simple} are both impressive methods in terms of reducing false positive rate.
Another edge-based method based on Steerable filter is used in many research papers such as in \cite{satzoda2014drive},\cite{sivaraman2013integrated}. A Steerable filter is convolved with an input image to select the maximum response in the direction of lane marking. This method has good effects when road markings are clearly painted and totally smooth. However, Steerable filter is not adapted to heavy traffic where the orientation of lane markings is not always dominant of all directions.

\subsubsection{Colours-Based Methods}
These methods are quite efficient for unstructured roads, or rural roads without clear lane boundaries. However, unlike edge-based methods, colour-based methods are not widely used by researchers. Colour information has its own drawbacks as it is influenced by lightening. Therefore, Detection based on colour does not work well for night scenes or heavy and complex traffic.
In \cite{sotelo2004color} used a colour-based method in the HIS color space by computing the cylindrical distribution of colour features. Lane detection based on colour methods on gray level is used in \cite{chiu2005lane}. 

\subsubsection{Hybrid Information Based Methods}
Hybrid information is usually composed of edge information and colour information. This type of feature extraction scheme usually combines width, length, location (coordinates of pixels) of lanes with gray level and brightness intensity of groups of points, which makes the extraction results better.
The example of using hybrid information is in \cite{schreiber2005single}, lane boundaries are deemed as a combination of four lines, which are left and right boundaries of left and right lane marking, respectively. lane marking is treated as a region with a certain thickness (space between left and right boundaries). The grading criterion for the hypothesis is first calculated separately for the left and right line pairs, and then the two are combined together. The grading scheme are based on both edge and gray-level information, which are mean support-size of a pair, mean perpendicular gradient of a pair, similarity of a pair, mean continuity of a pair and mean lateral accuracy.

\subsection{ Machine Learning Based Lane Detection Algorithms}
In spite of using conventional image processing based methods to detect lane markings, some investigators focus on detecting lane marking using novel machine learning and deep learning methods. In the last decade Deep Learning became one of the hottest topics and trended research areas due to the development of deep network theories, parallel computing techniques, and large scale data.
Many deep learning algorithms show great advantages in computer vision tasks and the detection and recognition performance increases dramatically compared to conventional approaches. The convolution neural network (CNN) is one of the most popular approaches used for the object recognition research. CNN provides some impressive properties such as high detection accuracy, automatic feature learning, and end-to-end recognition. Recently, some researchers have successfully applied CNN and other deep learning techniques to lane detection. It was reported that by using CNN model, the lane detection accuracy increased dramatically from 80\% to 90\% compared with traditional image processing methods \cite{he2016lane},\cite{xing2018advances}.

\section{Lane Tracking}
To enable the following of lane marking over time, a tracking stage is usually incorporated. The two most common tracking techniques used in lane position detection systems are Kalman filtering \cite{deng2013real}, \cite{satzoda2013vision} and Particle filtering \cite{kim2006realtime}, \cite{danescu2009probabilistic}. In these systems, feature extraction and position tracking are often combined into a closed-loop feedback system in which the tracked lane position is defined as an initial estimation of the location and orientation of the extracted features. This stage has the ability to reduce the detection of the misleading parts by comparing the current detection with previous frames and then improve the detection accuracy.

\subsection{Kalman Filter}
Previous work has been done on tracking the parameters of Hough transform by using Kalman Filter as in \cite{deng2013real} and \cite{satzoda2013vision}. Usually, as the output of standard Hough transform, ($\rho$,  $\theta$)  is used to initialize the Kalman filter. With proper setting of state transition matrix, the state vector can be updated, which consists of  ($\rho$,  $\theta$) and its derivative.
In spite of the advantages of Kalman filter in terms of real time performance, it is unable to reject outliers that cause failure of tracking, which in turn decreases the accuracy of detection. To alleviate these problems, an extended Kalman filter (EKF) for road tracking was introduced in \cite{tian2006single}. EKF is designed for the tracking stage of a non-linear dynamic systems, such as lane detection in complex road environment where the noise might pose as a non-Gaussian distribution.

\subsection{Particle Filter}
Particle filter is another reliable option for lane tracking. It has been used in \cite{kim2006realtime} and \cite{danescu2009probabilistic}.  In \cite{kim2006realtime} proposed a method, which combines lane border hypotheses obtained by RANSAC with hypotheses from a particle filter. In \cite{danescu2009probabilistic} takes more information into account, which are horizontal and vertical curvature, lateral offset and lane width. In terms of visual cues for robust lane tracking, \cite{danescu2009probabilistic} uses curbs and road edges as the cues, while \cite{apostoloff2004vision} uses more than those, which consists of lane marking, road edge, road and non-road colour and road width. Usually, more reliable cues make the lane tracking with particle filter performs well in more complex situations.\\

In \cite{danescu2009probabilistic} with a comparison between kalman fiter and particle filter pointed out that, particle filter and Kalman filter have their own advantages and disadvantages. Unlike a Kalman filter, a particle filter does not need initialization or measurement validation before updating the state matrix. Particles can evolve by themselves, and it might have the chance to cluster around the best lane estimation. However, the premise is the tracking system must be well designed and the measurement cues are relevant, which requires the system must know how to validate a lane candidate.

\section{Conclusion}
Lane detection is a fundamental aspect of most current Advanced Driver Assistance Systems. 
Several intelligent transportation system applications, such as automatic steering systems, lane departure warning systems, lane change warning, adaptive cruise control and lane keeping assistance systems, depend on obtaining accurate estimates of the vehicle’s trajectory with respect to the lane the vehicle is driving in.
The critical challenge that lane detection systems are faced with is the demand for high reliability and the diverse working conditions.
In this paper we have studied the different approaches of lane detection and tracking system. We focused on preprocessing block and  investigated the details of different methods.
The purpose of preprocessing stage is to enhance the input image in order to increase the likelihood of the successful delivery of areas with useful information to subsequent stages.
With proper implementation and appropriate selection of preprocessing methods, real time lane detector can be implemented with proper accuracy.

%Nearly half portion of the image contains information of road lanes and other half is the image of horizon, which is not useful for lane detection purpose.We have also studied the various challenges faced by the lane detection system. 


%\section*{Acknowledgment}






%\begin{thebibliography}{1}


%\end{thebibliography}

%\bibliographystyle{IEEEtran}
%\bibliography{Reference}

\begin{thebibliography}{00}

\bibitem{date2017vision} P. V. Date and V. Gaikwad, “Vision based lane detection and departure
warning system,” in 2017 International Conference on Signal Processing
and Communication (ICSPC). IEEE, 2017, pp. 240–245.
\bibitem{xing2018advances} Y. Xing, C. Lv, L. Chen, H. Wang, H. Wang, D. Cao, E. Velenis, and F.-Y. Wang, “Advances in vision-based lane detection: algorithms, integration, assessment, and perspectives on acp-based parallel vision,” IEEE/CAA Journal of Automatica Sinica, vol. 5, no. 3, pp. 645–661, 2018.
\bibitem{shirke2017study} S. Shirke and C. Rajabhushanam, “A study of lane detection techniques and lane departure system,” in 2017 International Conference on Algorithms, Methodology, Models and Applications in Emerging Technologies (ICAMMAET). IEEE, 2017, pp. 1–4.
\bibitem{shihavuddin2008road} A. Shihavuddin, K. Ahmed, M. S. Munir, and K. R. Ahmed, “Road
boundary detection by a remote vehicle using radon transform for path
map generation of an unknown area,” International Journal of Computer
Science and Network Security, vol. 8, no. 8, pp. 64–69, 2008.
      \bibitem{truong2008new} Q.-B. Truong and B.-R. Lee, “New lane detection algorithm for
autonomous vehicles using computer vision,” in 2008 International
Conference on Control, Automation and Systems. IEEE, 2008, pp.
1208–1213.
      \bibitem{truong2008lane} Q. B. Truong, B. R. Lee, N. G. Heo, Y. J. Yum, and J. G. Kim, “Lane
boundaries detection algorithm using vector lane concept,” in 2008 10th
International Conference on Control, Automation, Robotics and Vision.
IEEE, 2008, pp. 2319–2325.
      \bibitem{zheng2008automatic}B. Zheng, B. Tian, J. Duan, and D. Gao, “Automatic detection technique
of preceding lane and vehicle,” in 2008 IEEE International Conference.
on Automation and Logistics. IEEE, 2008, pp. 1370–1375.
      \bibitem{yu2008lane} B. Yu, W. Zhang, and Y. Cai, “A lane departure warning system based on
machine vision,” in 2008 IEEE Pacific-Asia Workshop on Computational
Intelligence and Industrial Application, vol. 1. IEEE, 2008, pp. 197–
201.
      \bibitem{hota2009simple} R. N. Hota, S. Syed, S. Bandyopadhyay, and P. R. Krishna, “A simple
and efficient lane detection using clustering and weighted regression.”
in COMAD. Citeseer, 2009.
      \bibitem{bao2008lane} T. Q. Bao, L. G. Ryong, H. N. Geon, Y. Y. Jin, and K. J. Gook,
“Lane boundaries detection algorithm using vector lane concept,” in
International Conference on Control, Automation, Robotics and Vision
(ICARCV’08), 2008.
      \bibitem{he2003color} Y. He, H. Wang, and B. Zhang, “Color based road detection in
urban traffic scenes,” in Proceedings of the 2003 IEEE International
Conference on Intelligent Transportation Systems, vol. 1. IEEE, 2003,
pp. 730–735.
      \bibitem{kong2009vanishing} H. Kong, J.-Y. Audibert, and J. Ponce, “Vanishing point detection for
road detection,” in 2009 IEEE Conference on Computer Vision and
Pattern Recognition. IEEE, 2009, pp. 96–103.

\bibitem{rasmussen2008roadcompass} C. Rasmussen, “Roadcompass: following rural roads with vision+ ladar
using vanishing point tracking,” Autonomous Robots, vol. 25, no. 3, pp.
205–229, 2008.
      \bibitem{hua2006fast} L. Hua-jun, G. Zhi-bo, L. Jian-feng, and Y. Jing-yu, “A fast method for
vanishing point estimation and tracking and its application in road images,” in 2006 6th International Conference on ITS Telecommunications.
IEEE, 2006, pp. 106–109.
      \bibitem{matessi1999vanishing}  A. Matessi and L. Lombardi, “Vanishing point detection in the hough
transform space,” in European Conference on Parallel Processing.
Springer, 1999, pp. 987–994.
\bibitem{moghadam2011fast} P. Moghadam, J. A. Starzyk, and W. S. Wijesoma, “Fast vanishing-point
detection in unstructured environments,” IEEE Transactions on Image
Processing, vol. 21, no. 1, pp. 425–430, 2011.
\bibitem{nieto2011road} M. Nieto, J. A. Laborda, and L. Salgado, “Road environment modeling
using robust perspective analysis and recursive bayesian segmentation,”
Machine Vision and Applications, vol. 22, no. 6, pp. 927–945, 2011.

      \bibitem{bellino2005lane} M. Bellino, Y. L. De Meneses, P. Ryser, and J. Jacot, “Lane detection
algorithm for an onboard camera,” in Photonics in the Automobile, vol.
5663. International Society for Optics and Photonics, 2005, pp. 102–
111.
      \bibitem{nedevschi20043d} S. Nedevschi, R. Schmidt, T. Graf, R. Danescu, D. Frentiu, T. Marita,
F. Oniga, and C. Pocol, “3d lane detection system based on stereovision,”
in Proceedings. The 7th International IEEE Conference on Intelligent
Transportation Systems (IEEE Cat. No. 04TH8749). IEEE, 2004, pp.
161–166.
      \bibitem{deng2013real} J. Deng and Y. Han, “A real-time system of lane detection and tracking
based on optimized ransac b-spline fitting,” in Proceedings of the 2013
Research in Adaptive and Convergent Systems. ACM, 2013, pp. 157–164.
\bibitem{sivaraman2013integrated}  S. Sivaraman and M. M. Trivedi, “Integrated lane and vehicle detection,
localization, and tracking: A synergistic approach,” IEEE Transactions
on Intelligent Transportation Systems, vol. 14, no. 2, pp. 906–917, 2013.
\bibitem{li2013sensor} Q. Li, L. Chen, M. Li, S.-L. Shaw, and A. Nuchter, “A sensor-fusion ¨
drivable-region and lane-detection system for autonomous vehicle navigation in challenging road scenarios,” IEEE Transactions on Vehicular
Technology, vol. 63, no. 2, pp. 540–555, 2013.
\bibitem{shen2013multi} Y. Shen, J. Dang, E. Ren, and T. Lei, “A multi-structure elements based
lane recognition algorithm,” Przeglkad Elektrotechniczny, vol. 89, pp.
206–210, 2013.

\bibitem{mccall2006video} J. C. McCall and M. M. Trivedi, “Video-based lane estimation and
tracking for driver assistance: survey, system, and evaluation,” 2006.
\bibitem{wu2009dynamic} B.-F. Wu, C.-T. Lin, and Y.-L. Chen, “Dynamic calibration and occlusion
handling algorithms for lane tracking,” IEEE Transactions on Industrial
Electronics, vol. 56, no. 5, pp. 1757–1773, 2009.
\bibitem{satzoda2014drive} R. K. Satzoda and M. M. Trivedi, “Drive analysis using vehicle dynamics
and vision-based lane semantics,” IEEE Transactions on Intelligent
Transportation Systems, vol. 16, no. 1, pp. 9–18, 2014.
      \bibitem{li2004lane} Q. Li, N. Zheng, and H. Cheng, “Lane boundary detection using an
adaptive randomized hough transform,” in Fifth World Congress on
Intelligent Control and Automation (IEEE Cat. No. 04EX788), vol. 5.
IEEE, 2004, pp. 4084–4088.
      \bibitem{lipski2008fast} C. Lipski, B. Scholz, K. Berger, C. Linz, T. Stich, and M. Magnor, “A
fast and robust approach to lane marking detection and lane tracking,” in
2008 IEEE Southwest Symposium on Image Analysis and Interpretation.
IEEE, 2008, pp. 57–60.
     \bibitem{sun2006hsi} T.-Y. Sun, S.-J. Tsai, and V. Chan, “Hsi color model based lane-marking
detection,” in 2006 IEEE Intelligent Transportation Systems Conference.
IEEE, 2006, pp. 1168–1172.
      \bibitem{wen2008road} Q. Wen, Z. Yang, Y. Song, and P. Jia, “Road boundary detection in
complex urban environment based on low-resolution vision,” in 11th
Joint International Conference on Information Sciences. Atlantis Press,
2008.
 \bibitem{sun2011automatic} H. Sun, C. Wang, and N. El-Sheimy, “Automatic traffic lane detection
for mobile mapping systems,” in 2011 International Workshop on MultiPlatform/Multi-Sensor Remote Sensing and Mapping. IEEE, 2011, pp.
1–5.
\bibitem{cualain2012automotive}  D. Cualain, C. Hughes, M. Glavin, and E. Jones, “Automotive standardsgrade lane departure warning system,” IET Intelligent Transport Systems,
vol. 6, no. 1, pp. 44–57, 2012.
\bibitem{sotelo2004color} M. A. Sotelo, F. J. Rodriguez, L. Magdalena, L. M. Bergasa, and
L. Boquete, “A color vision-based lane tracking system for autonomous
driving on unmarked roads,” Autonomous Robots, vol. 16, no. 1, pp.
95–116, 2004.
      \bibitem{chiu2005lane} K.-Y. Chiu and S.-F. Lin, “Lane detection using color-based segmentation,” in IEEE Proceedings. Intelligent Vehicles Symposium, 2005.
IEEE, 2005, pp. 706–711.
      \bibitem{schreiber2005single} D. Schreiber, B. Alefs, and M. Clabian, “Single camera lane detection
and tracking,” in Proceedings. 2005 IEEE Intelligent Transportation
Systems, 2005. IEEE, 2005, pp. 302–307.
      \bibitem{he2016lane} B. He, R. Ai, Y. Yan, and X. Lang, “Lane marking detection based
on convolution neural network from point clouds,” in 2016 IEEE 19th
International Conference on Intelligent Transportation Systems (ITSC).
IEEE, 2016, pp. 2475–2480.
      \bibitem{satzoda2013vision} R. Satzoda and M. Trivedi, “Vision-based lane analysis: Exploration
of issues and approaches for embedded realization,” in Proceedings
of the IEEE Conference on Computer Vision and Pattern Recognition
Workshops, 2013, pp. 604–609.
      \bibitem{kim2006realtime} Z. Kim, “Realtime lane tracking of curved local road,” in 2006 IEEE
Intelligent Transportation Systems Conference. IEEE, 2006, pp. 1149–
1155.
\bibitem{danescu2009probabilistic} R. Danescu and S. Nedevschi, “Probabilistic lane tracking in difficult
road scenarios using stereovision,” IEEE Transactions on Intelligent
Transportation Systems, vol. 10, no. 2, pp. 272–282, 2009.

      \bibitem{tian2006single} M. Tian, F. Liu, and Z. Hu, “Single camera 3d lane detection and
tracking based on ekf for urban intelligent vehicle,” in 2006 IEEE
International conference on vehicular electronics and safety. IEEE,
2006, pp. 413–418.
\bibitem{apostoloff2004vision} N. Apostoloff and A. Zelinsky, “Vision in and out of vehicles: Integrated
driver and road scene monitoring,” The International Journal of Robotics
Research, vol. 23, no. 4-5, pp. 513–538, 2004.













\end{thebibliography}

\end{document}